\documentclass[letterpaper]{article} 
\usepackage{aaai2026}  
\usepackage{times}  
\usepackage{helvet}  
\usepackage{courier}  
\usepackage[hyphens]{url}  
\usepackage{graphicx} 
\usepackage{natbib}  
\usepackage{caption} 
\usepackage{algorithm}
\usepackage{algorithmic}

\usepackage{inconsolata}
\usepackage{amsmath,amssymb,amsfonts}
\usepackage{subfigure}
\usepackage{textcomp}
\usepackage{xcolor}

\usepackage{enumitem}
\usepackage{bm}
\usepackage{multirow}
\usepackage{booktabs}
\usepackage{graphics}
\usepackage{keyval}
\usepackage{trig}
\usepackage{threeparttable}
\usepackage{marvosym}

\usepackage{newtxmath}
\usepackage{tabularx}
\usepackage{colortbl}

\usepackage{newfloat}
\usepackage{listings}
\DeclareCaptionStyle{ruled}{labelfont=normalfont,labelsep=colon,strut=off} 
\floatstyle{ruled}
\newfloat{listing}{tb}{lst}{}
\floatname{listing}{Listing}
\title{Failure-Informed Image Self-Augmentation for Multimodal Large Language Model Self-Improvement}
\author{
    Chunyang Jiang\textsuperscript{\rm 1}\equalcontrib, Pingping Zhang\textsuperscript{\rm 2}\equalcontrib, Yuzhi Zhao\textsuperscript{\rm 2}, Wenao Ma\textsuperscript{\rm 3}, Zhijian Hou\textsuperscript{\rm 2}, Mengyang Wu\textsuperscript{\rm 3}, Yiyang Cai\textsuperscript{\rm 1}, Senkang Hu\textsuperscript{\rm 2}, Sitong Cheng\textsuperscript{\rm 1}, Chi-Min Chan\textsuperscript{\rm 1}, Wei Xue\textsuperscript{\rm 1}, Yike Guo\textsuperscript{\rm 1}
}
\affiliations{
    \textsuperscript{\rm 1}Hong Kong University of Science and Technology\\
    \textsuperscript{\rm 2}City University of Hong Kong\\
    \textsuperscript{\rm 3}Chinese University of Hong Kong\\
    rubickjiang@gmail.com
}

\usepackage{bibentry}

\begin{document}

\maketitle

\begin{abstract}
Multimodal large language models (MLLMs) have achieved remarkable performance across vision-language tasks, but their progress depends heavily on large-scale, high-quality multimodal data that are costly to annotate. Self-augmentation offers a promising alternative by enabling models to expand their own training data without external supervision. However, existing MLLM self-augmentation methods are largely text-centric, while image augmentation remains underexplored and typically relies on generic or handcrafted transformations that are weakly aligned with the model’s actual incapability. We propose Failure-informed Image Self-Augmentation (\textbf{FISA}), a framework for MLLM self-improvement that constructs augmented images from the model’s own failure cases. Our method generates visually challenging yet answer-preserving image complications, verifies their utility through self-examination, and applies dual fidelity filtering to avoid semantic distortion. Experiments on visual question answering benchmarks show that the proposed method consistently improves performance across both in-distribution and out-of-distribution settings. 
Further experiments validate the compatibility of FISA with existing textual self-augmentation approaches, the superior data efficiency of the synthesized samples over generic image augmentation baselines, and the practical effectiveness of the proposed filtering strategy.
\end{abstract}

\section{Introduction}
Multimodal large language models (MLLMs) have recently achieved remarkable progress across a wide range of vision-language tasks~\citep{glm-v,qwen3-vl,phi-4}. Their rapidly improving capabilities are largely driven by the availability of large-scale, high-quality multimodal data~\citep{mllm-data-survey}. However, acquiring such complex and high-fidelity data typically requires reliable human annotations, which makes it expensive, time-consuming, and inherently limited in scale~\citep{run-out-of-data}. Therefore, reducing the dependence on manually curated data has emerged as an important research focus for further advancing multimodal large language models.

In response to this challenge, data augmentation-driven self-improvement (self-augmentation) methods have recently attracted increasing attention. Inspired by the well-established strategies for text-only large language models~\citep{self-instruct,self-alignment}, current MLLM self-augmentation methods typically start with a seed dataset and exploit the model’s own capabilities to synthesize new multimodal samples~\citep{enhancing-cognition,LLM-data-annotation-survey, ECD}, which are then filtered through unsupervised or self-supervised procedures. This paradigm allows the training set to be expanded with less dependence on costly external supervision and subsequently realizes MLLM self-improvement.

Despite the growing progress, existing self-augmentation approaches for MLLMs exhibit a pronounced modality imbalance. On the one hand, a substantial body of work~\citep{first-sft,genixer,oasis,enhancing-cognition} focuses primarily on augmenting the textual part of multimodal datasets, increasing the diversity and complexity of textual instructions and reasoning thoughts, by adapting methods originally developed for text-only LLMs~\citep{self-instruct}.
On the other hand, self-augmentation on the images remains relatively underexplored and often overly generic. Some existing unsupervised image augmentation practices adopt elementary transformations such as resizing, cropping, or rotation, to create counterparts for original images~\citep{enhancing-VLM,sherlock,ssl-r1}. Others depend on handcrafted, intuitive operations derived from domain-specific prior knowledge, such as augmenting geometric figures with auxiliary elements~\citep{theorem-validated,mathcanvas}. However, these approaches are either too trivial to actually improve semantic diversity or too specialized to generalize across different domains. More fundamentally, their synthesis strategies are largely capability-agnostic and intuition-driven, rather than grounded in the model’s actual failure modes, thereby offering the model marginal challenge and limited utilization of augmentation data.

To bridge this gap, we propose a \textbf{F}ailure-informed \textbf{I}mage \textbf{S}elf-\textbf{A}ugmentation (\textbf{FISA}) framework for MLLM self-improvement. Moving beyond generic image synthesis, our central idea is to instead generate augmentation images that are tailored to expose the model’s visual understanding incapability, and thereby are more efficient for self-training.
Concretely, we guide the MLLM through a three-stage instruction template to explore and propose complications of seed images, and employ self-examination to verify their utility by examining whether they genuinely expose the model's failure cases in multimodal understanding. Meanwhile, we constrain the complication process to preserve the necessary visual information underlying the original answers, so that no additional annotation effort is needed. A dual fidelity filtering strategy is further applied to discriminate samples that violate this constraint by introducing semantic distortion. The augmented samples that pass both utility and fidelity filters are ultimately used to expand the seed dataset, enabling more data-efficient MLLM self-improvement.

We evaluate our method on general visual question-answering (VQA) tasks and demonstrate its consistent effectiveness in improving the model's performance across various datasets, including both in-distribution and out-of-distribution settings. Further experiments show that our image self-augmentation strategy is orthogonal to existing textual instruction self-augmentation methods, without introducing interference. We also demonstrate the superior data-efficiency of failure-informed image augmentation compared to generic baselines, with the utility of our proposed filtering strategies further validated through ablation studies. In summary, this work highlights the contribution of failure-informed image augmentation in MLLM self-improvement and proposes a practical unsupervised framework to generate and filter such data-efficient augmentation samples.

\section{Related Work}
\subsection{Image Augmentation for MLLMs}
To address the growing demand for large-scale, high-quality training data in MLLMs, many image synthesis methods have been explored. AuditDM~\citep{AuditDM} adopts an adversarial training framework to construct a dedicated image generator for the target model, with guidance by discrepancy signals between the target model and stronger teacher models. Synthetic-VL~\citep{R1-SyntheticVL} leverages the in-context learning capability of MLLMs to iteratively refine their generation strategy based on previously generated examples, fostering diversity along multiple dimensions, with several strong supervisor models employed to ensure generation quality. ECD~\citep{ECD} focuses specifically on chart understanding, decoupling chart data generation from rendering functions. By formulating chart image generation as modifications to underlying data and plotting code, ECD utilizes GPT-4o to produce large quantities of diverse chart image samples.

While these works demonstrate the feasibility of generating scalable high-quality synthetic image data for MLLM training, they rely heavily on the supervision from external powerful models. A smaller line of work has investigated image synthesis strategies that do not depend on external supervision. However, such methods are often confined to specific domains. For example, MathCanvas~\citep{mathcanvas} and TR-CoT~\citep{theorem-validated} generate and augment geometric images through geometry reasoning engines, increasing image diversity and complexity via fixed rules such as adding auxiliary lines. While effective in their target domains, these approaches are difficult to generalize to broader multimodal tasks.

\subsection{Self-Improvement of MLLMs}
MLLM self-improvement primarily studies how to improve model capabilities in the absence of external supervision. Substantial prior work focuses on constructing pseudo-labels for unlabeled training data, broadly falling into self-supervised and unsupervised approaches. Self-supervised methods exploit the model's own capabilities as explicit supervision signals. For example, CSR~\citep{CSR} uses \textit{self-judging}~\citep{self-judge} to produce reward signals, and R3V~\citep{R3V} employs \textit{self-reflection}~\citep{self-reflection} to supervise intermediate reasoning processes. Unsupervised methods instead derive training signals without direct model introspection. EvoQuality~\citep{EvoQuality}, for instance, leverages the majority-voting~\citep{self-consistency} guide mechanism to build ranking signals, which then guide image quality assessment training.

Orthogonal to pseudo-labeling is a line of research that uses the model's own generative ability to synthesize training data, \textit{i.e.}, self-augmentation. Following the established augmentation paradigm~\citep{self-instruct} in text-only LLMs, several methods explore diversifying the textual elements of multimodal training samples, such as task instructions~\citep{first-sft} and reasoning chains-of-thought~\citep{enhancing-cognition}. Genixer~\citep{genixer} further specializes this process by training a dedicated instruction generator, improving the specificity and quality of the synthesized data. Oasis~\citep{oasis} instead takes a prompt-free approach, allowing the model to generate instructions without explicit prior guidance and thereby avoiding biases introduced by humans.

Compared with the growing body of work on textual self-generation, self-augmentation on images remains underexplored. Conventional image augmentation techniques~\citep{image-aug-survey}, such as rotation, blurring, and noise injection, are too generic for multimodal understanding tasks and may introduce semantic distortion. As a result, existing self-improvement methods often incorporate such unsupervised image perturbations only as degraded samples, rather than as a source of genuinely informative augmented training data. For example, some methods use them as the dis-preferred samples to construct preference pairs for DPO training~\citep{enhancing-VLM, sherlock}, while others employ them as inputs for contrastive learning to enforce consistency in visual semantic recognition~\citep{ssl-r1}.
\section{Methodology}
The overall framework of FISA consists of three stages: failure-oriented image complication, utility and fidelity filtering, and augmented training with self-explanations. Starting from a seed dataset, we first construct more challenging yet answer-preserving image augmentations. We then filter the generated samples to ensure that they both expose model failure cases and preserve coherent semantics. The retained samples are finally merged with the seed data and used for self-training with self-explanations.

\begin{figure}[t]
  \centering
  \includegraphics[width=0.95\columnwidth]{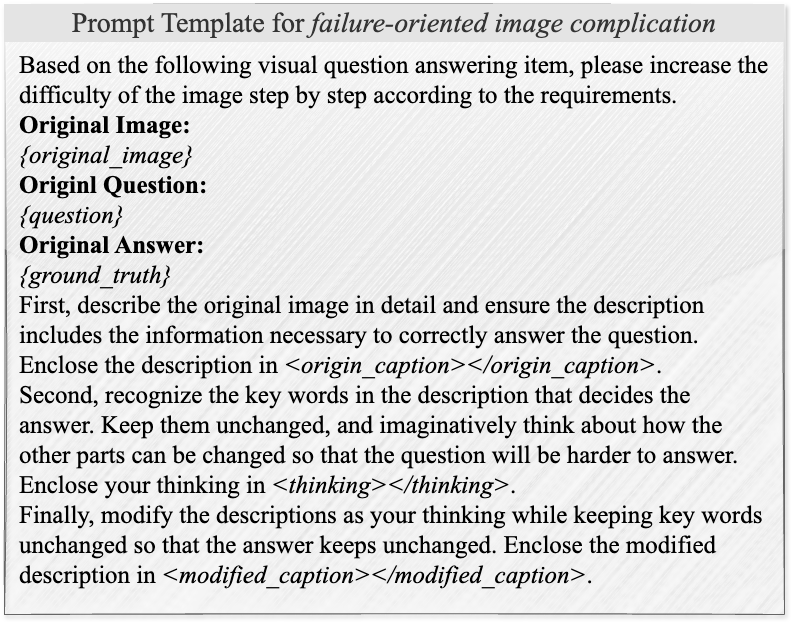}
  \caption{Instruction template for failure-oriented image complication.}
  \label{fig:instruction}
\end{figure}
\subsection{Failure-Oriented Image Complication}
Existing generic image augmentation methods, such as flipping, jittering, and rotation, are limited to surface-level perturbations, making them less effective for augmenting MLLM training~\citep{enhancing-VLM, sherlock}. Our goal, instead, is failure-oriented image complication, which increases visual difficulty through semantically coherent modifications that are more likely to reveal the target model’s failure cases.

A key challenge, however, is that meaningful complication requires flexible image modification, which may also distort the answer-critical information in the original image. Such distortion would require additional annotation, undermining the goal of self-improvement with minimal external supervision. We therefore seek an answer-preserving complication process that increases visual difficulty without changing the necessary information required to answer the original question.

To achieve this, we design a three-stage reasoning template that guides the model to produce failure-oriented yet answer-preserving image complication. As illustrated in Figure~\ref{fig:instruction}, the model is required to generate three components in sequence: \textit{original caption} $\mathcal{C}_o$, \textit{thinking} $\mathcal{T}$, and \textit{modified caption} $\mathcal{C}_m$. The original caption provides a detailed summary of the original image $\mathcal{I}_o$, including the information necessary for answering the question $\mathcal{Q}$. In the thinking stage, the model identifies how to make the image more challenging while keeping answer-critical information unchanged. In the final stage, the modified caption is deducted, describing the intended complicated image while explicitly retaining the information required to preserve the
original answer. Finally, the modified caption is fed into a pretrained text-to-image model to synthesize the corresponding modified image $\mathcal{I}_m$.

In our design, the complication process is carried out entirely in text before final image generation, which offers several advantages. First, preserving answer-critical information is substantially easier in textual descriptions than in direct image editing. Second, text-level complication promotes semantically coherent modifications rather than arbitrary low-level perturbations or noise, making the resulting samples more suitable for supervised training. Finally, the modified caption provides a natural reference for evaluating the fidelity of the generated image, which we will discuss in the next section.

\begin{table*}
    \centering
    \resizebox{0.9\textwidth}{!}{
    \begin{tabular}{lcccccccc}
    \toprule
    \multicolumn{2}{c}{} & A-OKVQA & SEEDBench & AI2D & MME & MMStar & MMBench & Average \\
    \midrule
    \multirow{4}{*}{Qwen3-VL-2B} & Vanilla & 74.76 & 72.60 & 64.25 & \textbf{77.25} & 46.20 & 80.92 & 69.33 \\
    & \cellcolor{gray!20}Vanilla+\textbf{FISA} & \cellcolor{gray!20}75.72 (\textcolor{red}{$\uparrow$}) & \cellcolor{gray!20}\textbf{73.10} (\textcolor{red}{$\uparrow$}) & \cellcolor{gray!20}66.00 (\textcolor{red}{$\uparrow$}) & \cellcolor{gray!20}76.24 (\textcolor{green!70!black}{$\downarrow$}) & \cellcolor{gray!20}50.73 (\textcolor{red}{$\uparrow$}) & \cellcolor{gray!20}81.20 (\textcolor{red}{$\uparrow$}) & \cellcolor{gray!20}70.50 (\textcolor{red}{$\uparrow$}) \\
    \cmidrule(r){2-9}
    & Vanilla+SI & \textbf{75.98} & 72.30 & 66.21 & 76.62 & 51.07 & 81.40 & 70.60 \\
    & \cellcolor{gray!20}Vanilla+SI+\textbf{FISA} & \cellcolor{gray!20}75.37 (\textcolor{green!70!black}{$\downarrow$}) & \cellcolor{gray!20}72.34 (\textcolor{red}{$\uparrow$}) & \cellcolor{gray!20}\textbf{66.45} (\textcolor{red}{$\uparrow$}) & \cellcolor{gray!20}77.17 (\textcolor{red}{$\uparrow$}) & \cellcolor{gray!20}\textbf{51.40} (\textcolor{red}{$\uparrow$}) & \cellcolor{gray!20}\textbf{81.68} (\textcolor{red}{$\uparrow$}) & \cellcolor{gray!20}\textbf{70.73} (\textcolor{red}{$\uparrow$}) \\
    \midrule
    \multirow{4}{*}{Gemma3-4B} & Vanilla & 78.43 & 69.40 & 68.04 & 79.74 & 44.20 & 80.32 & 70.02 \\
    & \cellcolor{gray!20}Vanilla+\textbf{FISA} & \cellcolor{gray!20}\textbf{79.04} (\textcolor{red}{$\uparrow$}) & \cellcolor{gray!20}\textbf{69.50} (\textcolor{red}{$\uparrow$}) & \cellcolor{gray!20}68.98 (\textcolor{red}{$\uparrow$}) & \cellcolor{gray!20}\textbf{79.91} (\textcolor{red}{$\uparrow$}) & \cellcolor{gray!20}\textbf{45.67} (\textcolor{red}{$\uparrow$}) & \cellcolor{gray!20}\textbf{80.46} (\textcolor{red}{$\uparrow$}) & \cellcolor{gray!20}\textbf{70.59} (\textcolor{red}{$\uparrow$}) \\
    \cmidrule(r){2-9}
    & Vanilla+SI & 77.47 & 68.60 & 68.62 & 78.85 & 43.20 & 80.32 & 69.51 \\
    & \cellcolor{gray!20}Vanilla+SI+\textbf{FISA} & \cellcolor{gray!20}77.73 (\textcolor{red}{$\uparrow$}) & \cellcolor{gray!20}68.80 (\textcolor{red}{$\uparrow$}) & \cellcolor{gray!20}\textbf{69.70} (\textcolor{red}{$\uparrow$}) & \cellcolor{gray!20}79.57 (\textcolor{red}{$\uparrow$}) & \cellcolor{gray!20}44.07 (\textcolor{red}{$\uparrow$}) & \cellcolor{gray!20}79.95 (\textcolor{green!70!black}{$\downarrow$}) & \cellcolor{gray!20}69.97 (\textcolor{red}{$\uparrow$}) \\
    \midrule
    \multirow{4}{*}{LLaVA1.5-7B} & Vanilla & 75.11 & 63.90 & 45.40 & 56.74 & 30.67 & 70.32 & 57.02 \\
    & \cellcolor{gray!20}Vanilla+\textbf{FISA} & \cellcolor{gray!20}75.37 (\textcolor{red}{$\uparrow$}) & \cellcolor{gray!20}63.10 (\textcolor{green!70!black}{$\downarrow$}) & \cellcolor{gray!20}45.95 (\textcolor{red}{$\uparrow$}) & \cellcolor{gray!20}63.65 (\textcolor{red}{$\uparrow$}) & \cellcolor{gray!20}31.87 (\textcolor{red}{$\uparrow$}) & \cellcolor{gray!20}70.34 (\textcolor{red}{$\uparrow$}) & \cellcolor{gray!20}58.38 (\textcolor{red}{$\uparrow$}) \\
    \cmidrule(r){2-9}
    & Vanilla+SI & \textbf{76.24} & 64.30 & 45.82 & 67.61 & 31.60 & 70.52 & 59.35 \\
    & \cellcolor{gray!20}Vanilla+SI+\textbf{FISA} & \cellcolor{gray!20}75.72 (\textcolor{green!70!black}{$\downarrow$}) & \cellcolor{gray!20}\textbf{64.50} (\textcolor{red}{$\uparrow$}) & \cellcolor{gray!20}\textbf{46.37} (\textcolor{red}{$\uparrow$}) & \cellcolor{gray!20}\textbf{68.66} (\textcolor{red}{$\uparrow$}) & \cellcolor{gray!20}\textbf{32.47} (\textcolor{red}{$\uparrow$}) & \cellcolor{gray!20}\textbf{70.82} (\textcolor{red}{$\uparrow$}) & \cellcolor{gray!20}\textbf{59.76} (\textcolor{red}{$\uparrow$}) \\
    \bottomrule
    \end{tabular}
    }
    \caption{Effectiveness of FISA in improving MLLM training performance. Two comparison settings are included: adding FISA to the seed dataset (\textbf{Vanilla} \textit{vs.} \textbf{Vanilla+FISA}), and further adding FISA after text augmentation with Self‑Instruct (\textbf{Vanilla+SI} \textit{vs.} \textbf{Vanilla+SI+FISA}).}
    \label{tab:main}
\end{table*}

\subsection{Utility and Fidelity Filtering}
We do not use all images generated in the previous stage for training. Instead, we filter every candidate from two complementary perspectives: \textbf{utility}, which measures whether the sample genuinely exposes a failure case of the model, and \textbf{fidelity}, which measures whether the generation preserves the answer-critical information. Since our image generation process involves two generative steps, first producing the modified caption and then synthesizing the image from it, the fidelity checking is further decomposed into text-level and image-level components accordingly.

We first define the utility checking signal. For each generated sample, we construct a testing triplet consisting of the modified image $\mathcal{I}_m$, the original question $\mathcal{Q}$, and the ground-truth answer $a_g$. We then feed the image-question pair $\langle \mathcal{Q}, \mathcal{I}_m \rangle$ into the model and obtain its prediction $a_u$ via greedy decoding. The utility checking signal $S_u$ is
\begin{equation}
    S_u = \mathbb{I}(a_u \neq a_g)
\end{equation}
Here, $S_u=1$ indicates that the generated image successfully induces failure of the model.

Still, utility checking alone is insufficient, since an incorrect answer may arise not only from increased difficulty but also from semantic distortion introduced during generation. We therefore perform two additional fidelity checks to verify that the generated sample still preserves the information required for the original answer.

We first conduct text-level fidelity checking on the modified caption $\mathcal{C}_m$. Specifically, we ask the model to answer the original question $\mathcal{Q}$ from only $\mathcal{C}_m$, and denote the predicted answer by $a_{vt}$. The text-level fidelity signal $S_{vt}$ is defined as
\begin{equation}
    S_{vt} = \mathbb{I}(a_{vt} = a_g)
\end{equation}
The intuition is that if $\mathcal{C}_m$ faithfully preserves the answer-critical information, then the correct answer should be literally recoverable from $\langle \mathcal{Q}, \mathcal{C}_m \rangle$. Therefore, $S_{vt}=1$ indicates that the modified caption remains faithful to the answer.

We next examine whether the generated image $\mathcal{I}_m$ faithfully follows the modified caption. Although automatic metrics such as CLIPScore~\citep{clipscore} are commonly used to assess text-to-image alignment, we empirically find them insufficient for verifying the fine-grained answer-critical details required in our setting. Instead, we adopt a rollout-based self-response criterion. For each pair $\langle \mathcal{I}_m, \mathcal{Q} \rangle$, we stochastically sample $N$ responses from the model and denote the response set by $\mathcal{A}_{vi}$. The image-level fidelity signal $S_{vi}$ is defined as
\begin{equation}
    S_{vi} = \mathbb{I}\{\sum_{a^i \in \mathcal{A}_{vi}}^N\mathbb{I}(a^i = a_g) \geq k\cdot N\}
\end{equation}
where $k \in [\frac{1}{N}, 1]$ controls the activation threshold. In other words, $S_{vi}=1$ if at least one sampled response answers the question correctly. The intuition is that if the generated image truly preserves the answer-critical information, the model should have a chance to recover the correct answer at least once. Admittedly, this criterion is imperfect, yet we show in experiments that, empirically, it provides a more effective filtering signal than other unsupervised alternatives.

Finally, the generated image is kept if and only if
\begin{equation}
    (S_u=1) \land (S_{vt} =1) \land (S_{vi} =1)
\end{equation}
\subsection{Augmented Training with Self-Explanations}
After generation and filtering, we obtain an augmented dataset composed of failure-informed images. We then combine it with the seed dataset for self-training. However, training only on image-question-answer triples may be suboptimal due to the distribution gap~\citep{self-distill}. To mitigate this issue, we further enrich the mixed training data with self-explanations.

Concretely, we prompt the model with the image, question, and ground-truth answer to generate an explanation. Each final training instance thereby consists of four components: image, question, explanation, and ground-truth.

\section{Experiments}
\subsection{Setup}
\noindent\textbf{Datasets}. We construct the seed dataset by sampling 1,500 instances from SEEDBench~\citep{SEEDBench} and 1,500 instances from A-OKVQA~\citep{A-OKVQA}. We then evaluate the model performance across six benchmarks, including SEEDBench, A-OKVQA, MME~\citep{mme}, MMBench~\citep{MMBench}, MMStar~\citep{mmstar}, and AI2D~\citep{AI2D}, which together cover both in-distribution and out-of-distribution evaluations.

\noindent\textbf{Models}. We evaluate our method on three prevailing MLLMs, including Qwen3-VL-2B~\citep{qwen3-vl}, Gemma3-4B~\citep{gemma3}, and LLaVA1.5-7B~\citep{Llava}, covering different model architectures and parameter scales.

\noindent\textbf{Implementation details}. All model generation in experiments is performed by the vLLM engine~\citep{vllm}. Regarding sampling settings, stochastic sampling is applied when generating modified captions and computing $S_{vi}$, with temperature set to $0.7$ and top-p set to $0.9$; the number of rollouts is $32$ and $16$, respectively. Greedy decoding is used for computing $S_u$ and $S_{vt}$, generating self-explanations, and the final evaluation. For the activation threshold of $S_{vi}$, we set $k=\frac{1}{N}$ unless otherwise indicated. For training, all models are fine-tuned with full parameters using a learning rate of $1e-6$. We use Z-Image-Turbo as the text-to-image generator~\citep{z-image} for all settings. Training of Qwen and Gemma is conducted on 8×RTX 4090 GPUs, while LLaVA is trained on 2×H800 GPUs. The prompt templates for generating modified captions are shown in Figure~\ref{fig:instruction}. Other instruction templates, including computing $R_u$, $R_{vt}$, and $R_{vi}$, generating self-explanations, SFT training, and final evaluation, are shown in Appendix A.

\subsection{Main Results}
To evaluate whether FISA-generated data can genuinely improve MLLM training, we examine its effectiveness under two settings: directly augmenting the seed dataset with FISA (Vanilla \textit{vs.} Vanilla+FISA), and further augmenting self-instruct-augmented~\citep{self-instruct} seed data with FISA (Vanilla+SI \textit{vs.} Vanilla+SI+FISA). The evaluation is conducted on six benchmarks covering both in-distribution datasets (A-OKVQA and SEEDBench) and out-of-distribution datasets (AI2D, MME, MMStar, and MMBench). The results are shown in Table~\ref{tab:main}.

In the first setting, we compare training performance on the seed dataset alone against that on the mixture of the seed dataset and FISA-generated data. As the results show, augmenting the seed dataset with FISA improves the resulting performance in the majority of cases. This trend holds for both in-distribution and out-of-distribution evaluation. In the second setting, FISA is deployed on the self-instruct-augmented seed dataset, which reflects whether FISA remains effective when combined with existing textual augmentation. Similar to the first group of comparison, introducing FISA leads to further performance gains in most cases. These results suggest that, for multimodal understanding tasks, image-side augmentation and text-side augmentation constitute two largely orthogonal directions of improvement. Rather than interfering with each other, they are compatible and can work together to provide complementary benefits.

\subsection{Ablation on Failure-Oriented Image Complication}
\begin{table}
    \centering
    \resizebox{\columnwidth}{!}{
    \begin{tabular}{lcccccccc}
    \toprule
    \multicolumn{2}{c}{Datasets} & A-OKVQA & SEEDBench & AI2D & MME & MMStar & MMBench & Avg. \\
    \midrule
    \multirow{2}{*}{Qwen} & Cap2Aug & 75.11 & 72.80 & \textbf{66.52} & 74.81 & 49.80 & 81.01 & 70.01 \\
    \cmidrule(r){2-2} \cmidrule(r){3-9}
    & FISA & \textbf{75.72} & \textbf{73.10} & 66.00 & \textbf{76.24} & \textbf{50.73} & \textbf{81.20} & \textbf{70.50} \\
    \midrule
    \multirow{2}{*}{Gemma} & Cap2Aug & 77.38 & 68.50 & 68.46 & 79.70 & 43.33 & 80.25 & 69.60 \\
    \cmidrule(r){2-2} \cmidrule(r){3-9}
    & FISA & \textbf{79.04} & \textbf{69.50} & \textbf{68.98} & \textbf{79.91} & \textbf{45.67} & \textbf{80.46} & \textbf{70.59} \\
    \midrule
    \multirow{2}{*}{LLaVA} & Cap2Aug & 74.32 & 61.30 & \textbf{46.05} & 63.31 & 30.87 & 69.85 & 57.62 \\
    \cmidrule(r){2-2} \cmidrule(r){3-9}
    & FISA & \textbf{75.37} & \textbf{63.10} & 45.95 & \textbf{63.65} & \textbf{31.87} & \textbf{70.34} & \textbf{58.38} \\
    \bottomrule
    \end{tabular}
    }
    \caption{Comparison of Cap2Aug and FISA for image augmentation training under the same data volume.}
    \label{tab:self-summary}
\end{table}

\begin{figure}[t]
  \centering
  \includegraphics[width=0.95\columnwidth]{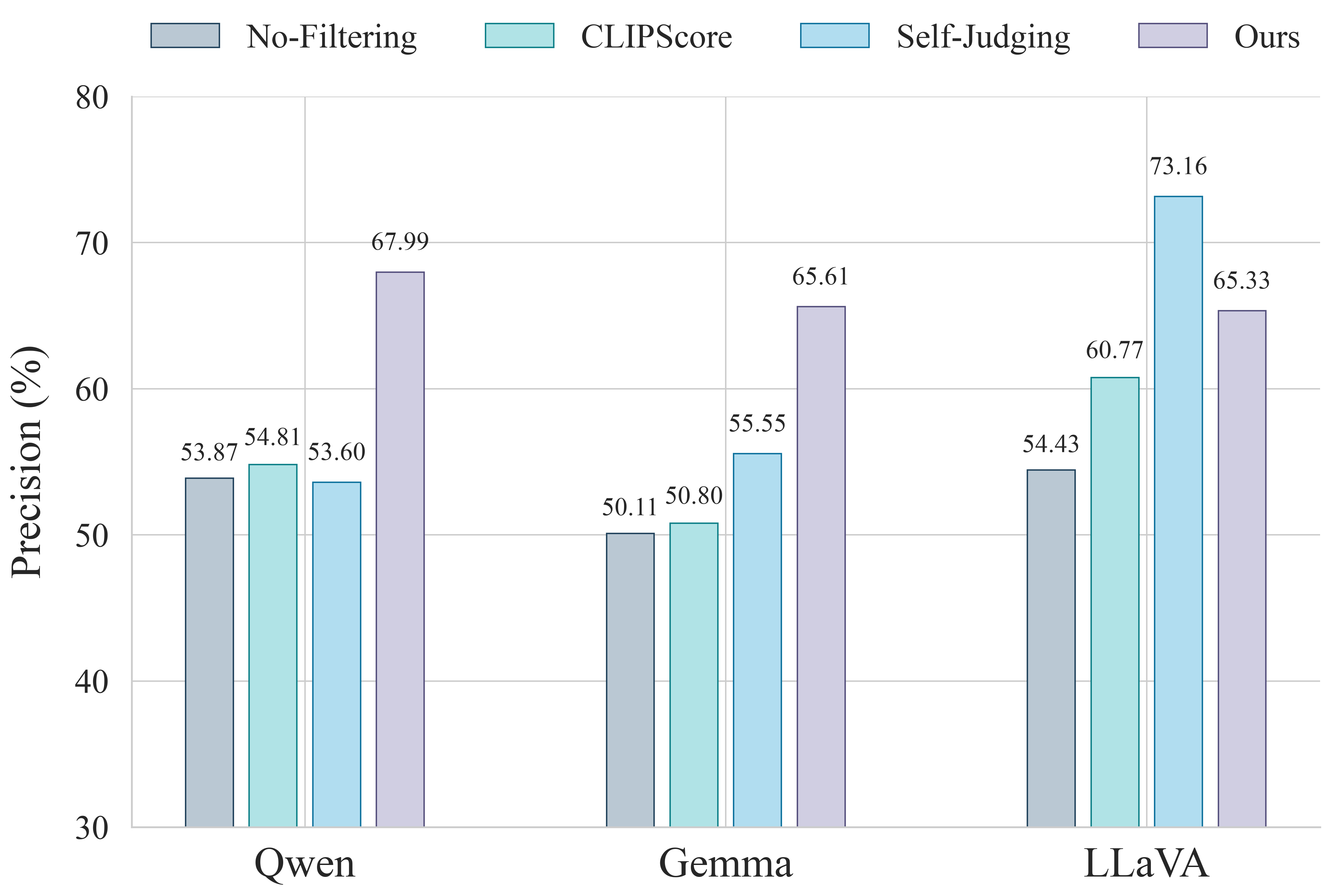}
  \caption{Comparison of our image‑level validity filtering strategy ($S_{vi}$-based) with alternative filtering methods (CLIPScore‑based and Self‑Judging‑based), evaluated by the precision of retained samples.}
  \label{fig:validity-ablation}
\end{figure}
In this section, we conduct ablation studies to evaluate whether the failure-oriented image complication approach actually holds a meaningful advantage over generic methods. For the baseline, since prior work~\citep{enhancing-VLM, sherlock} has shown that simple geometric transformations like translation and rotation tend to generate only degraded samples for MLLM training, we adopt a stronger augmentation baseline, Cap2Aug~\citep{Cap2Aug}, where the augmented image is synthesized from a textual description of the original image, yielding samples that are semantically more coherent and diverse. The core idea of Cap2Aug could be easily adapted to our self-improvement setting by letting the model self-summarize descriptions, and its only distinction from FISA is the absence of any failure-inducing operation in generating descriptions. This comparison allows us to assess whether explicitly guiding the model to actively discover failure modes improves the utilization of resulting augmented samples.
As the results presented in Table~\ref{tab:self-summary} show, within the same augmentation data budget, FISA consistently outperforms Cap2Aug across most cases, demonstrating superior data efficiency of the failure-oriented image augmentation for self-improvement.

\subsection{Ablation on Fidelity Filtering}
\label{sec:ablation}
In this section, we further conduct an ablation study on the fidelity filtering strategies. Since the text-level filtering ($S_{vt}$-based), which checks whether the modified caption $\mathcal{C}_m$ preserves the answer-critical information, is quite straightforward, we focus this study more on the image-level filtering ($S_{vi}$-based). Specifically, we compare our filtering based on $S_{vi}$, which measures image generation fidelity by whether at least one answer yields correct across multiple rollouts, against other feasible unsupervised alternatives designed for image fidelity.

We consider two baselines. The first is CLIPScore~\citep{clipscore}, a standard metric widely used to evaluate text-to-image generation fidelity. Specifically, we filter by retaining samples with the score above $0.99$. The second is Self-Judging~\citep{CSR, self-judge}, which leverages the MLLM's own capability to determine whether the generated image $\mathcal{I}_m$ is consistent with the modified caption $\mathcal{C}_m$. Concretely, we prompt the model to answer such verification $N$ (identical to $S_{vi}$) times and keep a sample if the majority of responses are positive. For fairness, all three methods are applied to the same candidate pool that has passed $S_{vt}$ filtering. Since manual evaluation entails an excessive workload, we instead use GPT-5.4~\citep{GPT-5} as the external judge to estimate the true precision of the retained sample set. The final results are shown in Figure~\ref{fig:validity-ablation}.

As results show, our $S_{vi}$-based filtering consistently improves the precision of the retained sample set from around 50\% to above 65\% across all three models. By contrast, CLIPScore filtering yields only a modest gain of about 5\% on LLaVA and is nearly ineffective on the other two models. Empirically, we found that most samples received very high and close scores regardless of whether they faithfully keep the answer-critical information, making discrimination almost impossible. This is likely because, in our setting, only fine-grained details that determine the answer matter, whereas CLIPScore is better suited to measuring coarse semantic alignment and is less sensitive to such subtle differences. Self-Judging exhibits the largest variance. It delivers the greatest improvement on LLaVA, yet causes degradation on Qwen. This suggests that the model's ability to act as a reliable judge varies substantially and is not necessarily correlated with its underlying capability. Overall, due to the more stable performance across different models, we adopt $S_{vi}$ as our final image-level fidelity filtering signal.

\begin{figure}[t]
  \centering
  \includegraphics[width=0.95\columnwidth]{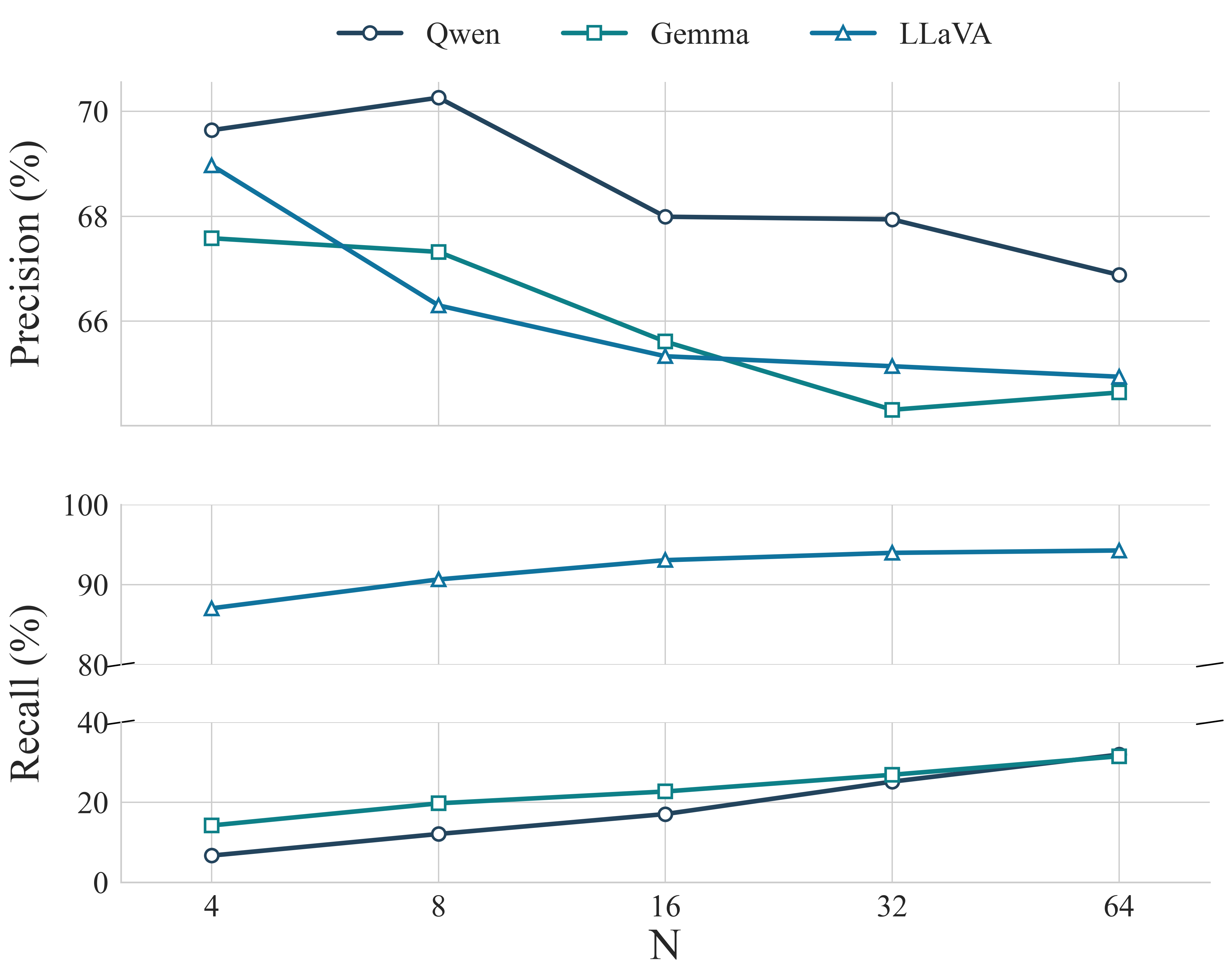}
  \caption{Precision and recall of retained samples from our image‑level validity filtering strategy with different rollout numbers $N$.}
  \label{fig:rollout-num}
\end{figure}
\begin{figure}[t]
  \centering
  \includegraphics[width=0.95\columnwidth]{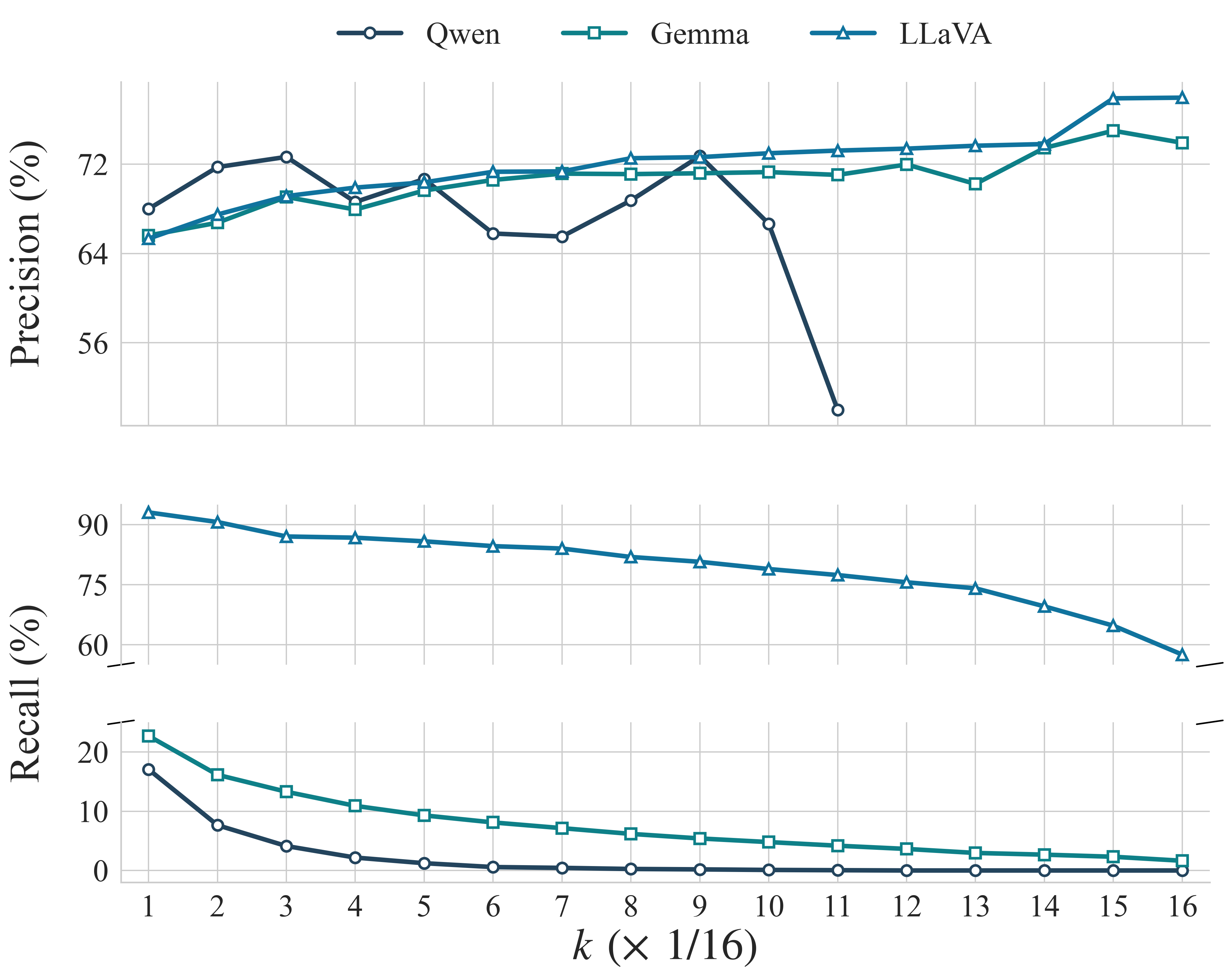}
  \caption{Precision and recall of retained samples from our image‑level validity filtering strategy with different activation thresholds $k$.}
  \label{fig:conf-rate}
\end{figure}
\subsection{Hyperparameter Study}
Important hyperparameters in FISA are primarily involved in computing $S_{vi}$, consisting of two key quantities: the number of sampling rollouts $N$ applied to each individual example, and the activation threshold $k$, which determines the minimum proportion of correct answers required to set $S_{vi} = 1$. In this section, we investigate how the values of these two parameters influence the filtering performance.

We first examine the effect of the number of sampling rollouts $N$ on filtering performance. Intuitively, a larger $N$ yields a greater diversity of model responses to each individual question, which simultaneously increases the probability of selecting true positive samples and admitting false positive samples. Consequently, the recall is expected to rise monotonically with $N$, while the change of precision depends on the relative rates at which true and false positives accumulate. Accordingly, we apply five rollout numbers, $N=4$, $N=8$, $N=16$, $N=32$, and $N=64$, across three models, measuring the precision and recall of $S_{vi}$-based filtering with GPT-5.4 as the judge. The results are presented in Figure~\ref{fig:rollout-num}. As shown, recall increases consistently with $N$ across all three models, in line with our hypothesis. Precision, while exhibiting minor local fluctuations, displays an overall downward trend as $N$ grows, suggesting that the rate of false positive accumulation outpaces that of true positives as the model is given more opportunities to respond. These results highlight a trade-off that, an excessively large $N$ degrades filtering precision, while an overly small $N$ risks discarding genuine positive samples. In practice, we adopt intermediate values such as $8$ or $16$ as a balance.

We then analyze the role of the activation threshold $k$ on filtering performance. In principle, increasing $k$ imposes a stricter retention criterion, as more correct responses are required for a sample to be retained. Consequently, both true positive and false positive samples are expected to decrease, leading to a corresponding decline in recall. The change in precision, however, remains uncertain, depending on the relative magnitudes of these two reductions. We show how precision and recall empirically vary with $k$ in Figure~\ref{fig:conf-rate}. As we can see, for both Gemma and LLaVA, precision exhibits an overall upward trend as $k$ increases, suggesting a greater reduction in false positives than that in true positives; Qwen, however, displays a more oscillatory pattern. Yet across all models, this improvement in precision comes at the expense of a substantial reduction in recall, which is particularly severe for Qwen. When $k$ increases from $\frac{1}{16}$ to $\frac{2}{16}$, recall drops by more than half, reflecting a sharp collapse in the total number of retained samples. And when $k \geq \frac{12}{16}$, no samples in Qwen are kept. The extremely limited number of retained samples also explains the pronounced fluctuations in Qwen’s precision. A similar phenomenon can be observed for Gemma, whereas LLaVA is relatively less affected. In addition to its detrimental effect on recall, increasing $k$ also entails an implicit cost. A higher proportion of correct responses generally implies that the retained samples are easier for the model. That is, the samples excluded as $k$ increases are likely to be more challenging, which conflicts with our motivation. Therefore, while a larger $k$ can improve the precision of the retained data, we ultimately adopt the smallest valid threshold $k=\frac{1}{N}$ in practice.

\begin{figure*}[t]
  \centering
  \label{fig:case-study}
  \includegraphics[width=0.95\textwidth]{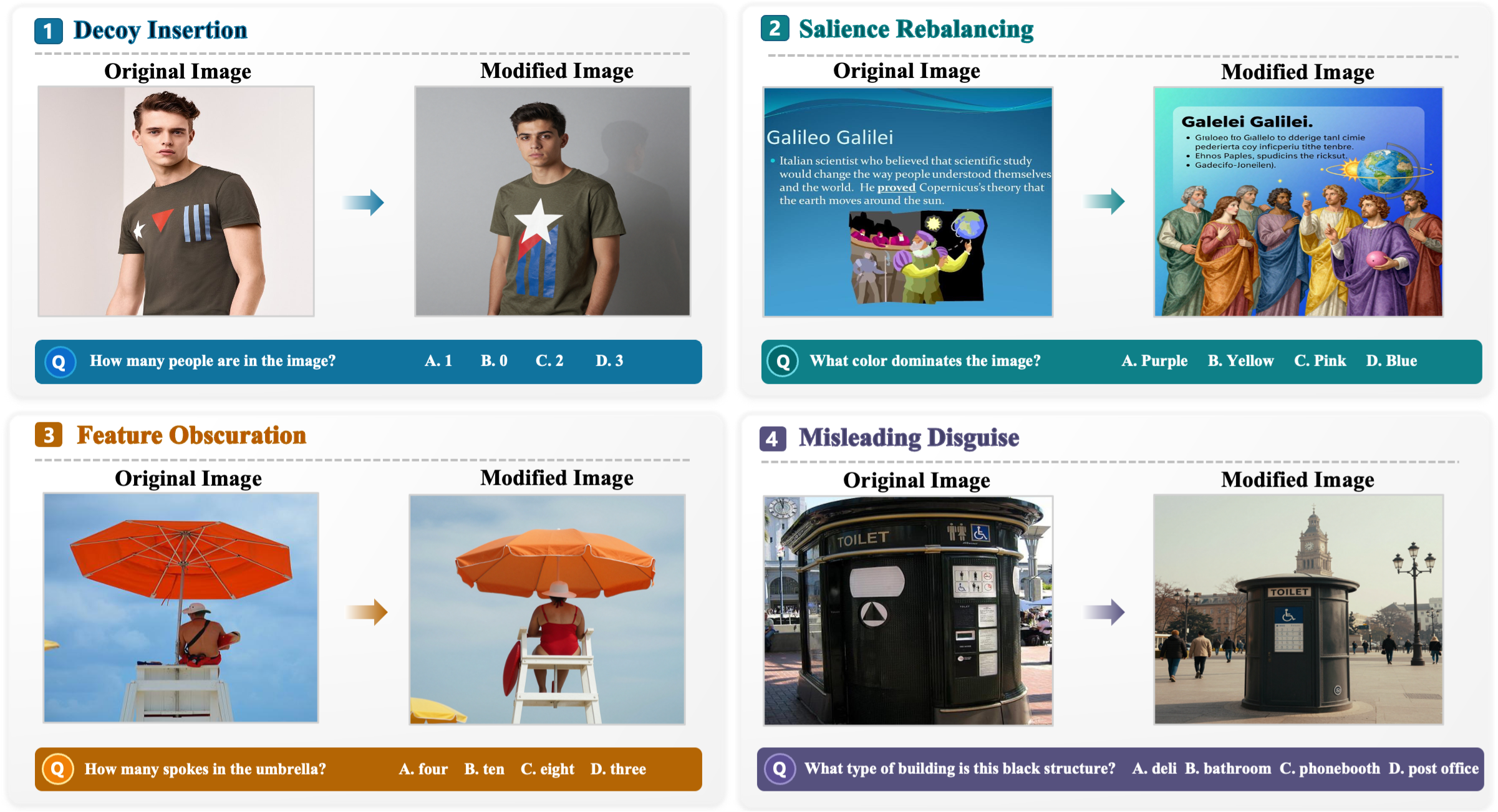}
  \caption{Examples illustrating four principal complication directions of FISA, including decoy insertion, feature obscuration, salience rebalancing, and misleading disguise.}
  \label{fig:case-study}
\end{figure*}

\subsection{Qualitative Study}
In this section, we qualitatively characterize the dimensions along which FISA complicates the original images. Through manual comparison between the FISA-generated images and their originals, we identify four principal directions of complication:
\begin{itemize}
    \item \textbf{Decoy Insertion}. Introduces spurious objects, figures, or visual elements into the image. These distractors either bear a close resemblance to the target element and are readily confounded with it, or share the same semantic category yet violate the state, action, or identity constraints specified in the question. The model is thereby compelled to identify the veridical candidate among multiple within-category alternatives, predisposing it to counting or judgment errors.

    \item \textbf{Feature Obscuration}. Attenuates the visibility, completeness, or canonical presentation of the target object's discriminative features through occlusion, shadowing, scaling reduction, or non-canonical viewpoints. Cues that are otherwise recognizable at a glance are thereby transformed into features requiring local-detail verification, inducing recognition or counting errors.
    
    \item \textbf{Salience Rebalancing}. Redistributes the salience distribution within the image, such as color composition and depth proportion, without altering its object inventory. The rebalancing is achieved by either attenuating the originally dominant key feature to a merely marginal advantage or elevating background distractors to a salience level nearly comparable to that of the main subject. The model is thus compelled to perform relative-magnitude judgments instead of glance-level dominance assessment, inducing judgment errors.
    
    \item \textbf{Misleading Disguise}. Modifies the target object's visual appearance to approximate an incorrect option in color, shape, or count; alternatively, homogenizes originally distinct objects into indistinguishable replicas, or disguises a single object as multiple. The model is thereby forced to bypass holistic appearance and rely on local details such as text, texture, or merchandise type, inducing category or counting errors.
    
\end{itemize}
For illustration, we exhibit concrete examples for each direction in Figure~\ref{fig:case-study}.

We have summarized the distribution of these four modes. Overall, Decoy Insertion and Feature Obscuration account for the largest proportions, together comprising nearly 70\% of the total. Salience Rebalancing constitutes the smallest share, at less than 10\%. When comparing the distributions across different models, both Qwen and Gemma exhibit basically consistent with the overall distribution, whereas LLaVA shows a notably higher proportion of Misleading Disguise, accompanied by lower proportions for the other three modes. Further details are provided in Appendix B.

\section{Conclusion}
In this work, we address the pronounced modality imbalance in MLLM self-augmentation, where image-side augmentation remains underexplored and is dominated by generic, capability-agnostic transformations. To bridge this gap, we propose Failure-informed Image Self-Augmentation (\textbf{FISA}), a framework that constructs augmented images from the model's own failure cases. FISA guides the model through a three-stage reasoning template to generate visually challenging yet answer-preserving image complications, retains the resulting candidates through a joint utility and dual fidelity filtering strategy, and finally merges them with the seed data for self-training with self-explanations. Experiments across three MLLMs and six benchmarks demonstrate that FISA consistently improves performance in both in-distribution and out-of-distribution settings, remains compatible with existing textual self-augmentation approaches, and achieves superior data efficiency over generic image augmentation baselines. Further ablation studies validate the effectiveness of failure-oriented complication and the practical utility of the proposed fidelity filtering. Overall, this work highlights the value of failure-informed image augmentation in MLLM self-improvement and offers a practical unsupervised framework for generating and filtering data-efficient augmentation samples.
\clearpage

\bibliography{aaai2026}
\clearpage

\appendix
\begin{figure}[h]
  \centering
  \includegraphics[width=0.95\columnwidth]{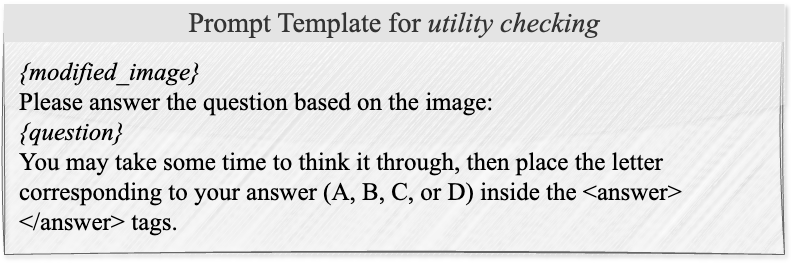}
  \caption{Shared instruction template for utility-checking, image-level fidelity checking, and performance evaluation.}
  \label{fig:utility-checking}
\end{figure}
\begin{figure}[h]
  \centering
  \includegraphics[width=0.95\columnwidth]{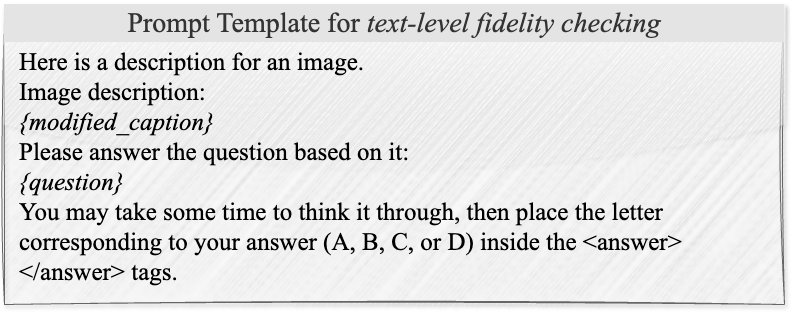}
  \caption{Instruction template for text-level fidelity checking.}
  \label{fig:fidelity-checking}
\end{figure}
\begin{figure}[h]
  \centering
  \includegraphics[width=0.95\columnwidth]{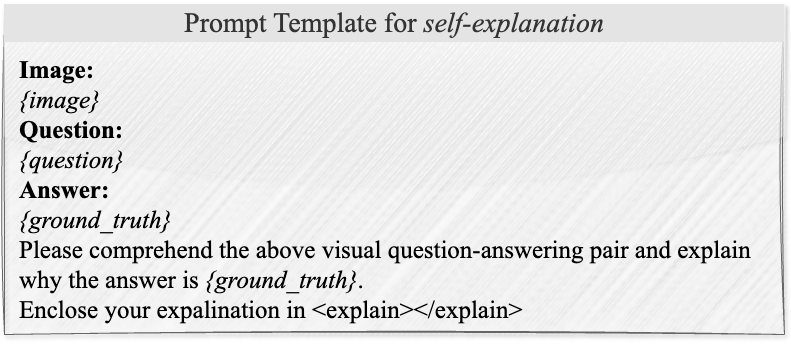}
  \caption{Instruction template for self-explanation.}
  \label{fig:explain}
\end{figure}

\section{Appendix}
\subsection{A Instruction Details}
\begin{figure}[h]
  \centering
  \includegraphics[width=0.95\columnwidth]{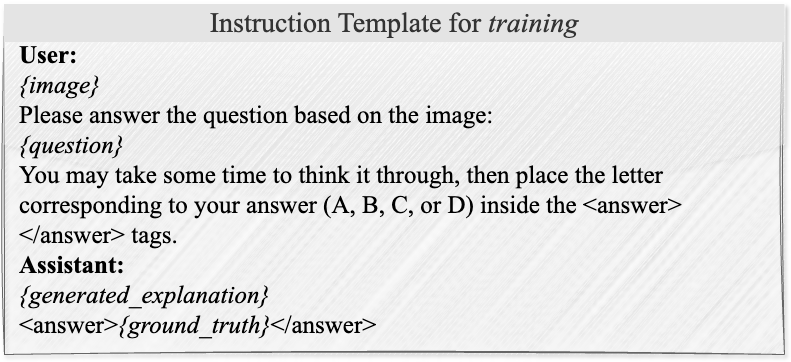}
  \caption{Instruction template for training.}
  \label{fig:sft}
\end{figure}

In this section, we show more detailed instruction templates, including those for computing $R_u$, $R_{vt}$, and $R_{vi}$, generating self-explanations, SFT training, and final evaluation. Figure~\ref{fig:utility-checking} shows the common template we use to prompt the model to answer questions based on the corresponding images. Computing $R_u$, $R_{vi}$, and the final evaluation all share this template, differing in decoding strategy. Figure~\ref{fig:fidelity-checking} shows how the model is prompted to answer questions based on image captions, which is used to compute text-level fidelity $R_{vt}$. Templates for generating self-explanations and SFT training are shown in Figure~\ref{fig:explain} and Figure~\ref{fig:sft}, respectively.

\subsection{B Qualitative Study}
\begin{figure}[t]
  \centering
\includegraphics[width=\columnwidth]{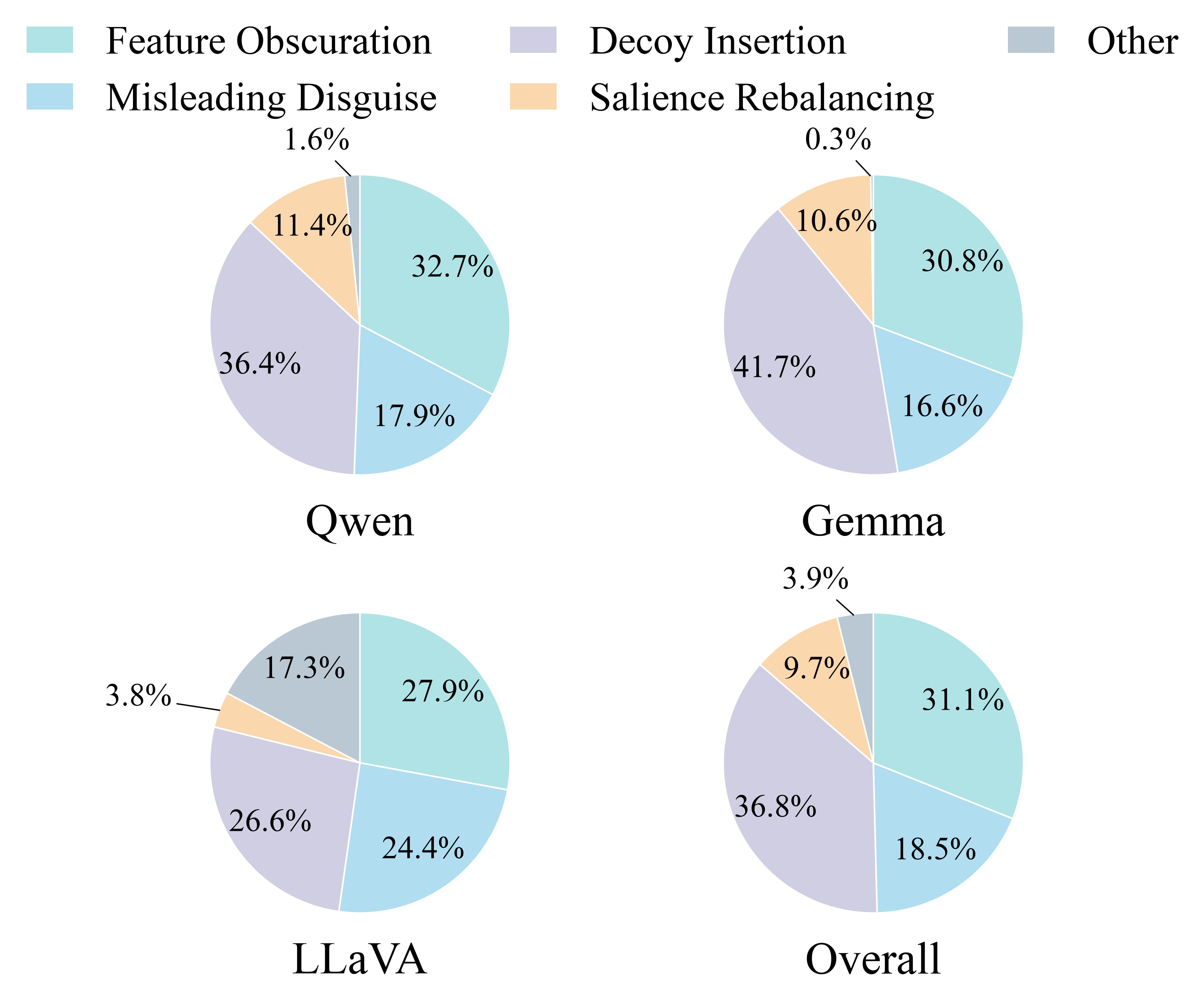}
  \caption{Distribution of FISA complication directions across different models.}
  \label{fig:pattern}
\end{figure}
\begin{figure*}[t]
  \centering
  \includegraphics[width=\textwidth]{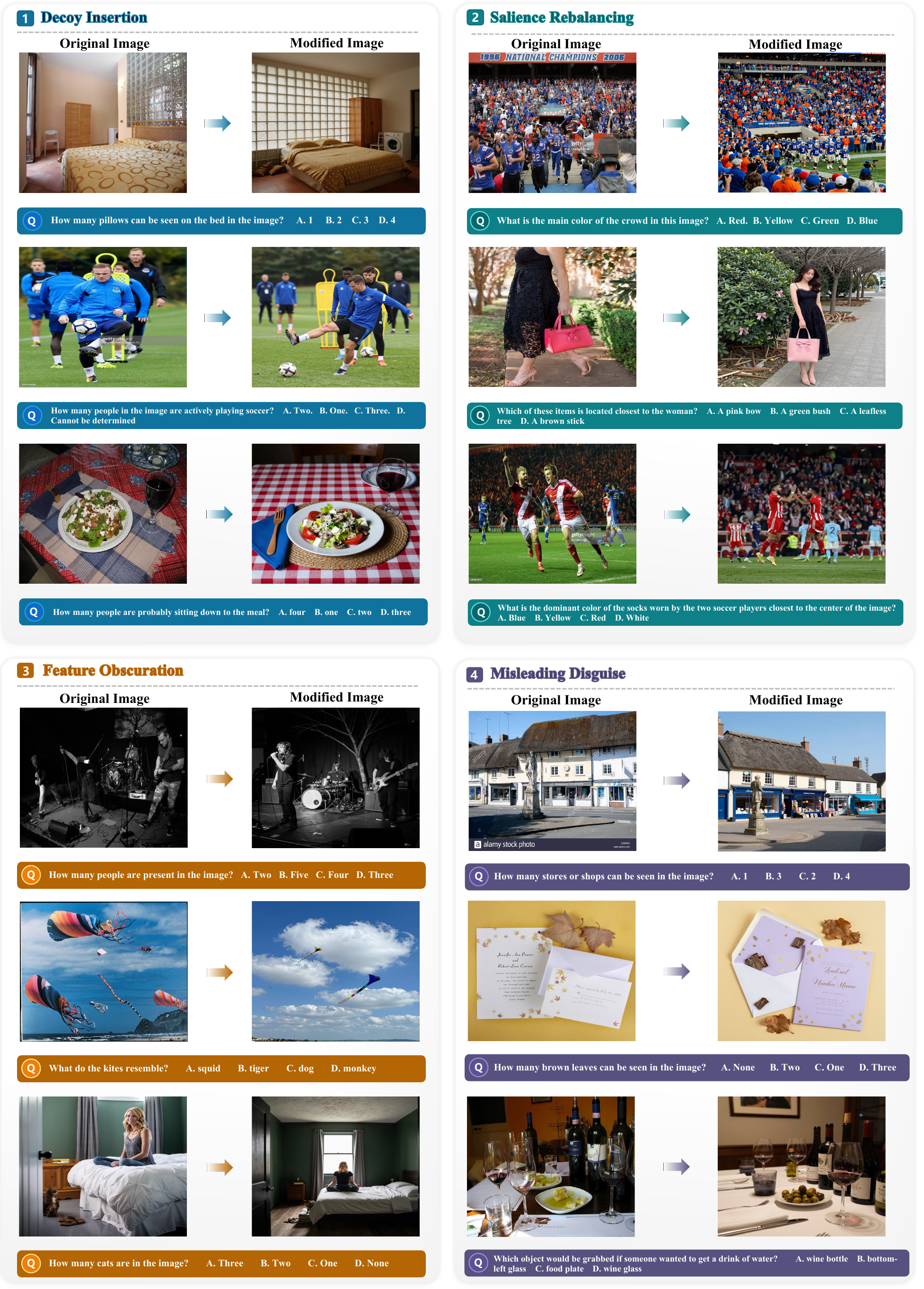}
  \caption{Examples illustrating four principal complication directions of FISA.}
  \label{fig:case-appendix}
\end{figure*}

We identify four principal directions along which FISA complicates the original images: \textit{Decoy Insertion}, \textit{Feature Obscuration}, \textit{Salience Rebalancing}, and \textbf{Misleading Disguise}. We then instruct GPT-5.4 to review all generated cases and assign each case to one of these four categories, with an additional \textit{Other} category reserved for cases that do not fit any of them. Figure~\ref{fig:pattern} presents the distribution of these patterns across different models. We also provide more representative examples illustrating each major direction in Figure~\ref{fig:case-appendix}.

\end{document}